\documentclass{article}

\usepackage{arxiv}

\usepackage[utf8]{inputenc} 
\usepackage[T1]{fontenc}    
\usepackage{hyperref}       
\usepackage{url}            
\usepackage{booktabs}       
\usepackage{amsfonts}       
\usepackage{nicefrac}       
\usepackage{microtype}      
\usepackage{xcolor}         
\usepackage{graphicx}
\usepackage{amsmath}
\usepackage{caption}
\usepackage{wrapfig}
\usepackage{float}
\usepackage{picinpar}

\title{Deep $k$-grouping: An Unsupervised Learning Framework for Combinatorial Optimization on Graphs and Hypergraphs}

\author{%
  Sen Bai\\
  Changchun University of Science\\
  and Technology, China \\
  \texttt{baisen@cust.edu.cn} \\
  \And
  Chunqi Yang \\
  Changchun University of Science\\
  and Technology, China \\
  \texttt{yangchunqi@mails.cust.edu.cn} \\
  \AND
  Xin Bai \\
  Huawei Technologies Co. Ltd \\
  China \\
  \texttt{baixinbs@163.com} \\
  \And
  Xin Zhang \\
  Changchun University of Science\\
  and Technology, China \\
  \texttt{zhangxin@cust.edu.cn} \\
  \And
  Zhengang Jiang \\
  Changchun University of Science\\
  and Technology, China \\
  \texttt{jiangzhengang@cust.edu.cn} \\
}

\begin{document}

\maketitle

\begin{abstract}

Along with AI computing shining in scientific discovery, its potential in the combinatorial optimization (CO) domain has also emerged in recent years.
Yet, existing unsupervised neural network solvers struggle to solve $k$-grouping problems (e.g., coloring, partitioning) on large-scale graphs and hypergraphs, due to limited computational frameworks. In this work, we propose Deep $k$-grouping, an unsupervised learning-based CO framework. Specifically, we contribute: \textbf{(\uppercase\expandafter{\romannumeral1})} Novel \textit{one-hot encoded polynomial unconstrained binary optimization} (OH-PUBO), a formulation for modeling $k$-grouping problems on graphs and hypergraphs (e.g., graph/hypergraph coloring and partitioning);
\textbf{(\uppercase\expandafter{\romannumeral2})} GPU-accelerated algorithms for large-scale $k$-grouping CO problems. Deep $k$-grouping employs the relaxation of large-scale OH-PUBO objectives as differentiable loss functions and trains to optimize them in an unsupervised manner.
To ensure scalability, it leverages GPU-accelerated algorithms to unify the training pipeline;
\textbf{(\uppercase\expandafter{\romannumeral3})} A Gini coefficient-based continuous relaxation annealing strategy to enforce discreteness of solutions while preventing convergence to local optima.
Experimental results demonstrate that Deep $k$-grouping outperforms existing neural network solvers and classical heuristics such as SCIP and Tabu.

\end{abstract}

\section{Introduction}
The majority of combinatorial optimization (CO) problems are NP-hard or NP-complete, making large-scale CO instances computationally intractable.
While problem-specific heuristics exist, a general approach is to formulate CO problems as integer programs (IP) and solve them with IP solvers.
However, IP solvers lack scalability for large-scale CO instances.
This has motivated recent research to explore end-to-end neural network solvers~\cite{ye2023gnn,tian2024neural,li2024power,cappart2023combinatorial} as a scalable alternative.

In the field of neural CO, unsupervised learning approaches~\cite{schuetz2022combinatorial,nazi2019deep,tsitsulin2023graph,ichikawa2024controlling,chen2024learning,wang2021neural} show promising prospects.
Unsupervised learning approaches eliminate the need for datasets of pre-solved problem instances.
Instead, they leverage CO objectives as their loss functions, optimizing these objectives by training the machine learning models.
However, neural networks typically operate in a differentiable parameter space, while CO requires discrete solution spaces.

Consequently, unsupervised neural network-based CO frameworks have two restrictions: \textbf{(\romannumeral1)} CO problems must be expressible in unconstrained objectives rather than IP problems composed of objectives and constraints; \textbf{(\romannumeral2)} They necessitate differentiable relaxations of discrete variables.
To this end, a line of unsupervised neural network solvers~\cite{schuetz2022combinatorial,ichikawa2024controlling,chen2024learning} formulate CO problems as \textit{quadratic unconstrained binary optimization} (QUBO) problems.
The QUBO problem is to optimize the following cost function:
\begin{equation}
\mathrm{min} \quad O_{\mathrm{QUBO}}=\sum_{i=1}^{n}\sum_{j=1}^{n}{Q_{ij}x_ix_j}=\mathbf{x}^TQ\mathbf{x}
\label{equ:QUBODef}
\end{equation}
where $\mathbf{x}=(x_1,x_2,\cdots,x_n)$ is an $n$-dimensional vector of binary decision variables $x_i\in \{0,1\}$.
The QUBO matrix $Q$ is either symmetric or in upper triangular form.
\begin{figure*}[t]
\centering
\includegraphics[width=1\columnwidth]{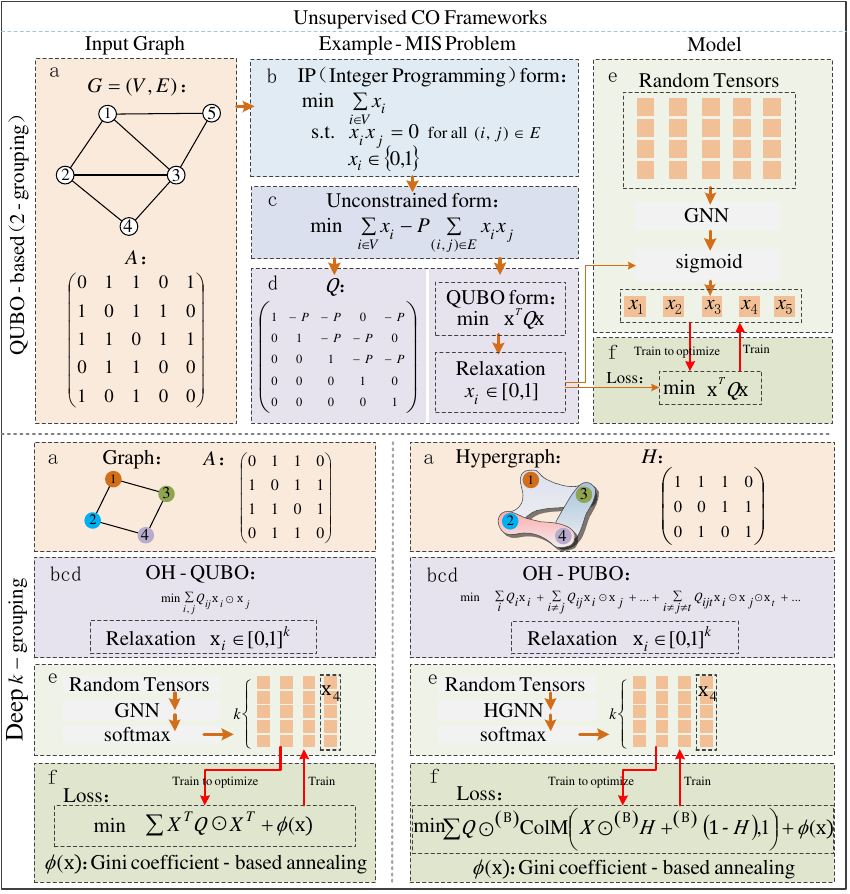}
\caption{Overview of unsupervised neural network-based CO frameworks, including QUBO-based neural network solvers and the proposed Deep $k$-grouping framework.}
\label{fig:framework}
\end{figure*}

QUBO aligns with unsupervised neural CO in three aspects: \textbf{(\romannumeral1)} Unconstrained objectives. It uses penalty method~\cite{glover2022quantum} to incorporate constraints into the objective function; \textbf{(\romannumeral2)} Well-suited activation functions. Sigmoid function provides a natural mechanism for generating the relaxation of binary variables $x_i\in[0,1]$; \textbf{(\romannumeral3)} GPU-accelerated loss functions with uniform formulation. $\mathbf{x}^TQ\mathbf{x}$ can be efficiently computed through deep learning frameworks like PyTorch via matrix operations.
The training pipeline for QUBO-based methods is illustrated in Fig.~\ref{fig:framework}, incorporating a case study of the maximum independent set (MIS) problem in Appendix~\ref{app:MISDef}.

Nevertheless, QUBO-based methods are restricted to $2$-grouping problems (MIS, max-cut, etc.).
When applied to $k$-grouping problems on graphs or hypergraphs, QUBO-based methods face challenges. \textbf{Challenge 1:} QUBO employs binary variables $x_i\in\{0,1\}$, which are compatible with only $2$-grouping tasks (such as distinguishing dominators and dominatees in the MIS problem); \textbf{Challenge 2:} Quadratic terms such as $x_ix_j$ cannot represent multi-variable correlations such as coloring conflicts within hyperedges. Thus, QUBO formulations fail to effectively model CO problems on hypergraphs.
These challenges significantly limit the scalability of unsupervised CO frameworks on numerous problems, such as graph/hypergraph coloring and partitioning.

An intuitive idea to address \textbf{Challenge 1} is to employ one-hot vectors instead of binary variables.
To tackle multi-variable correlations in \textbf{Challenge 2}, we generalize QUBO to \textit{polynomial unconstrained binary optimization} (PUBO). These solutions lead us to \textbf{Challenge 3:} Is there an efficient loss function with uniform formulation for learning-based $k$-grouping frameworks in the way of QUBO-based approaches?
To address these challenges, we present Deep $k$-grouping, an unsupervised neural network solver for solving CO problems. We contribute:

\textbf{1) OH-QUBO/OH-PUBO Formulations.} Novel \textit{one-hot encoded quadratic (or polynomial) unconstrained binary optimization} formulations for modeling $k$-grouping CO problems, encoding CO objectives and constraints via Hadamard products of one-hot vectors.
By leverging the differentiable softmax relaxation of one-hot variables, Deep $k$-grouping enables end-to-end unsupervised optimization.

\textbf{2) GPU-Accelerated Differentiable Optimization.}
A training pipeline enhancing the scalability of Deep $k$-grouping when applied to large-scale CO instances.

\textbf{3) Gini Coefficient-based Annealing}.
A  continuous relaxation annealing strategy serves a dual purpose: in the high-temperature phase, it broadly searches the solution space to escape local optima; in the low-temperature phase, it enforces the discreteness of solutions.

\textbf{4) Novel Cost Functions for Graph/Hypergraph Coloring/Partitioning}. An empirical study on these problems demonstrates the performance of Deep $k$-grouping.

\textbf{Notations.} For a $k$-grouping problem on a graph or hypergraph $G=(V,E)$, we have:

$\mathbf{x}_i$: the group assignment of a vertex $v_i$ is denoted by a $k$-dimensional one-hot vector $\mathbf{x}_i\in \{0,1\}^k$.

$X$: the group assignment matrix of all vertices $[\mathbf{x}_1,\mathbf{x}_2,\cdots,\mathbf{x}_{|V|}]$ is denoted by $X\in \mathbb{R}^{|V|\times k}$.

$\sum{\mathbf{x}}$ or $\sum{X}$: the sum of all elements of a vector $\mathbf{x}$ or a matrix $X$ is denoted as $\sum{\mathbf{x}}$ or $\sum{X}$.

$\phi(\mathbf{x})$: the Gini coefficient-based penalty term (refer to Eq.~\ref{equ:Gini}, Sec.~\ref{sec:annealing}).

\section{Method}
Deep $k$-grouping (Fig.~\ref{fig:framework}) solves $k$-grouping CO problems on graphs and hypergraphs in two stages: \textbf{(\romannumeral1)} formulating CO problems as unconstrained cost functions, including OH-QUBO or OH-PUBO (Sec.~\ref{sec:OHQUBO} and Sec.~\ref{sec:proglem_instances}), and 
\textbf{(\romannumeral2)} relaxing them into differentiable loss functions and training to solve them via neural network-based optimizer (Sec.~\ref{sec:NNarchitecture} and Sec.~\ref{sec:annealing}).


\subsection{Unconstrained Formulation of $k$-grouping CO Problems}
\label{sec:OHQUBO}
OH-QUBO and OH-PUBO employ one-hot vectors $\mathbf{x}_i\in \{0,1\}^k$ as decision variables to denote the group assignments.
The constraints within a set of vertices are represented via Hadamard products\footnote{Hadamard product is denoted by $\odot$. For notational consistency, $\mathbf{x}_i\odot\mathbf{x}_j$ is still described as quadratic-cost terms.} such as $\sum (\mathbf{x}_1 \odot \cdots \odot \mathbf{x}_m)=\begin{cases}
            1, & \text{if assigned the same group,} \\
            0, & \text{if assigned more than one group.}
        \end{cases}$

\textbf{Definition 1 (One-Hot Encoded QUBO or OH-QUBO).} Given a graph $G=(V,E)$, the cost function of an OH-QUBO problem on $G$ can be expressed as:
\begin{equation}
\mathrm{min} \quad O_{\mathrm{OH-QUBO}}=\sum_{i=1}^{|V|}\sum_{j=1}^{|V|}{Q_{ij}\mathbf{x}_i\odot\mathbf{x}_j}
\label{equ:OHQUBODefinition}
\end{equation}
Similar to QUBO, $\mathbf{x}_i\odot \mathbf{x}_i=\mathbf{x}_i$, thus linear terms can be omitted from Eq.~\ref{equ:OHQUBODefinition}.
The OH-QUBO matrix $Q$ is either symmetric or in upper triangular form.$\hfill\square$

\textbf{GPU-accelerated OH-QUBO Cost Function.} Eq.~\ref{equ:OHQUBODefinition} is equivalent to:
\begin{equation}
O_{\mathrm{OH-QUBO}}=\sum{X^TQ\odot X^T}
\label{equ:GPUOHQUBO}
\end{equation}
\textbf{Proof.} See Appendix~\ref{app:proof-Eq3}.$\hfill\square$

\textbf{Time Complexity Analysis.} For $X^T \in \mathbb{R}^{k \times |V|}$ and $Q \in \mathbb{R}^{|V| \times |V|}$, the operation $X^T Q \odot X^T$ has time complexity $O(k|V|^2)$. 
With GPU acceleration utilizing $P$ cores, this reduces to $O\left(\frac{k|V|^2}{P}\right)$ under perfect parallelization.

\textbf{Definition 2 (One-Hot Encoded PUBO or OH-PUBO).} The cost function of OH-PUBO problems can be expressed as:
\begin{equation}
\mathrm{min} \quad O_{\mathrm{OH-PUBO}}=\sum{(\sum_{i}{Q_i\mathbf{x}_i}+\sum_{i\neq j}{Q_{ij}\mathbf{x}_i\odot\mathbf{x}_j}+\sum_{i\neq j\neq t}{Q_{ijt}\mathbf{x}_i\odot\mathbf{x}_j\odot\mathbf{x}_t}+\cdots)}
\label{equ:OHPUBODef}
\end{equation}
\textbf{Definition 3 (Hypergraph).} A hypergraph $G=(V,E)$ is defined as a set of nodes $V=\{v_1,v_2,\cdots,v_{|V|}\}$ and a set of hyperedges $E=\{e_1,e_2,\cdots,e_{|E|}\}$, where each hyperedge $e_j\in E$ is a subset of $V$.
A hypergraph $G$ can be represented by an incidence matrix $H\in \{0,1\}^{|V|\times |E|}$, where $H_{ij}=1$ if $v_i\in e_j$, and $H_{ij}=0$ otherwise.$\hfill\square$

In hypergraph proper coloring, polynomial terms in Eq.~\ref{equ:OHPUBODef} such as $\mathbf{x}_1\odot\mathbf{x}_2\odot\mathbf{x}_3$ can be used to represent the color conflicts within a hyperedge $e_j=\{v_1,v_2,v_3\}$.
Our idea of GPU-accelerated OH-PUBO stems from the observation of one-to-one mapping correspondence between $O_\mathrm{OH-PUBO}$ in Eq.~\ref{equ:OHPUBODef} and a hypergraph $G=(V,E)$, where each polynomial term in $O_\mathrm{OH-PUBO}$ corresponds to a hyperedge $e_j$ in $E=\{e_1,e_2,\cdots,e_{|E|}\}$.

\textbf{GPU-accelerated OH-PUBO Cost Function.} Given a hypergraph $G=(V,E)$, where each hyperedge $e_j$ of $G$ corresponds to a (linear or polynomial) term in Eq.~\ref{equ:OHPUBODef}.
$H$ is the incidence matrix of $G$.
We define the PUBO matrix $Q\in \mathbb{R}^{1\times |E|}$ as $Q=[Q_{1},Q_{2},\cdots,Q_{|E|}]$, where $Q_j$ is the coefficient for each term in Eq.~\ref{equ:OHPUBODef} that corresponds hyperedge $e_j$, thus,
\begin{equation}
O_{\mathrm{OH-PUBO}}=\sum{Q\odot ^{(\mathrm{B})}{\mathrm{ColM}(X\odot ^{(\mathrm{B})}H+^{(\mathrm{B})}(1-H), 1)}}
\label{equ:OHPUBOTheorem}
\end{equation}
where $+^{(\mathrm{B})}$ or $\odot ^{(\mathrm{B})}$ denotes the element-wise addition or Hadamard product via the broadcasting mechanism.
$\mathrm{ColM}(M, 1)$ is the column-wise multiplication over the first dimension of matrix $M$.

\textbf{Proof.} See Appendix~\ref{app:proof-Eq5}.$\hfill\square$

\textbf{Time Complexity Analysis.} For $X \in \mathbb{R}^{|V| \times k}$, $Q \in \mathbb{R}^{1 \times |E|}$, and $H \in \mathbb{R}^{|V| \times |E|}$, the operation in Eq.~\ref{equ:OHPUBODef} has time complexity $O(k|V||E|)$.
GPU significantly speeds up the computation.
Element-wise operations such as Hadamard product are fully parallelizable.
Column-wise reduction such as sum or product over $|V|$ uses parallel reduction techniques, leading to time complexity $O(\mathrm{log}|V|)$.
Overall, the theoretical GPU time complexity is $O(\mathrm{log}|V|)$ under unrealistic core counts, or otherwise utilizing $P$ cores, the realistic GPU time can reduce to $O\left(\frac{k|V||E|}{P}\right)$.

\subsection{$k$-grouping Problem Instances}
\label{sec:proglem_instances}
Based on OH-QUBO and OH-PUBO, we propose novel cost functions for several $k$-grouping CO problems, including graph/hypergraph coloring and partitioning.
Graph and hypergraph coloring have numerous applications in areas such as image segmentation~\cite{tuan2021colorrl,gomez2007graph}, task scheduling~\cite{stollenwerk2020toward,darwish2021towards,ganguli2017study}, and resource allocation~\cite{wu2020coloring,huang2023hypergraph,wang2020uav,mukherjee2022grover}.
They refer to the assignment of minimal number of colors to vertices such that no edge or hyperedge is monochromatic.

\textbf{Graph Coloring.} Given a graph $G=(V,E)$, graph coloring problems can be formulated by optimizing the cost function as follows:
\begin{equation}
\mathrm{min} \quad O_{\mathrm{GC}}=\sum_{k=1}^{K_{\mathrm{max}}}{y_k}+\lambda_1\sum{\sum_{(v_i,v_j)\in E}{\mathbf{x}_i\odot\mathbf{x}_j}}+\lambda_2\sum_{i=1}^{|V|}{\sum_{k=1}^{K_{\mathrm{max}}}{x_{ik}(1-y_k)}}+\phi(\mathbf{x})
\label{equ:GC-OHQUBO}
\end{equation}
where $\lambda_1$,$\lambda_2>0$ are penalty parameters, $\mathbf{x}_i=(x_{i1},x_{i2},\cdots,x_{i{K_{\mathrm{max}}}})$, $\mathbf{x}_i\in \{0,1\}^{K_{\mathrm{max}}}$ are one-hot vectors that represent the color assignments, variables $y_k\in\{0,1\}$ denote whether the $k$-th color is used, and $K_{\mathrm{max}}$ is the predefined maximum value of chromatic number.
We minimize $\sum_{k=1}^{K_{\mathrm{max}}}{y_k}$ to optimize the chromatic number.
Meanwhile, two types of constraints must be satisfied: \textbf{(\romannumeral1)} $\sum_{(v_i,v_j)\in E}{\mathbf{x}_i\odot\mathbf{x}_j}$ ensures any pair of vertices within the same edge must be assigned different colors; \textbf{(\romannumeral2)} $\sum_{i=1}^{|V|}{\sum_{k=1}^{K_{\mathrm{max}}}{x_{ik}(1-y_k)}}$ ensures that $x_{ik}\leq y_k$, thus the $k$-th color should not be used by any vertex if $y_k=0$.
During training, $\lambda_1\sum_{(v_i,v_j)\in E}{\mathbf{x}_i\odot\mathbf{x}_j}$ can be implemented through GPU-accelerated OH-QUBO cost function in Eq.~\ref{equ:GPUOHQUBO}, where $Q_{ij}=1$ if $(v_i,v_j)\in E$, or otherwise $Q_{ij}=0$. Other terms in Eq.~\ref{equ:GC-OHQUBO} can be implemented through matrix operations.

\textbf{Hypergraph Strong Coloring.} Given a hypergraph $G=(V,E)$, the hypergraph strong coloring problem (all vertices in any given hyperedge have distinct colors) is to optimize the cost function:
\begin{equation}
\mathrm{min} \quad O_{\mathrm{HSC}}=\sum_{k=1}^{K_{\mathrm{max}}}{y_k}+\lambda_1\sum{\sum_{\stackrel{(v_i,v_j)\in e}{e\in E}}{\mathbf{x}_i\odot\mathbf{x}_j}}+\lambda_2\sum_{i=1}^{|V|}{\sum_{k=1}^{K_{\mathrm{max}}}{x_{ik}(1-y_k)}}+\phi(\mathbf{x})
\label{equ:HSC-OHQUBO}
\end{equation}
where $e$ are hyperedges in $E$.
Similarly, $\lambda_1\sum{\sum_{\stackrel{(v_i,v_j)\in e}{e\in E}}{\mathbf{x}_i\odot\mathbf{x}_j}}$ can be implemented through Eq.~\ref{equ:GPUOHQUBO}, where $Q_{ij}=1$ if $(v_i,v_j)\in e$ and $e\in E$, or otherwise $Q_{ij}=0$.

\textbf{Hypergraph Proper Coloring.} Given a hypergraph $G=(V,E)$, the hypergraph proper coloring problem (no hyperedge is monochromatic) is to optimize the cost function:
\begin{equation}
\mathrm{min} \ O_{\mathrm{HSC}}=\sum_{k=1}^{K_{\mathrm{max}}}{y_k}+\lambda_1\sum{\sum_{\stackrel{(v_{j1},\cdots,v_{j|e|})\in e}{e\in E}}{\mathbf{x}_{j1}\odot\cdots\odot\mathbf{x}_{j|e|}}}+\lambda_2\sum_{i=1}^{|V|}{\sum_{k=1}^{K_{\mathrm{max}}}{x_{ik}(1-y_k)}}+\phi(\mathbf{x})
\label{equ:HSC-OHQUBO}
\end{equation}
where $e$ are hyperedges in $E$.
Similarly, $\lambda_1\sum{\sum_{\stackrel{(v_{j1},\cdots,v_{j|e|})\in e}{e\in E}}{\mathbf{x}_{j1}\odot\cdots\odot\mathbf{x}_{j|e|}}}$ can be implemented through GPU-accelerated OH-PUBO cost functionin (Eq.~\ref{equ:OHPUBOTheorem}), where $Q_{i}=1$.

Graph and hypergraph partitioning serve as building blocks in numerous applications, including VLSI design~\cite{karypis1997multilevel,li2024mapart}, image segmentation~\cite{gomes2017stochastic,chai2024novel}, storage
sharding in distributed databases~\cite{kabiljo2017social,yang2018hepart}, and simulations of distributed quantum circuits~\cite{andres2019automated,gray2021hyper}.
They seek to partition the node set of a graph or hypergraph into multiple similarly-sized disjoint blocks while reducing the number of edges (hyperedges) that span different blocks.

\textbf{Graph Partitioning.} Given a graph $G=(V,E)$, the graph partitioning problem can be formulated by optimizing the following cost function:
\begin{equation}
\mathrm{min} \quad O_{\mathrm{GP}}=\alpha\sum{\sum_{(v_i,v_j)\in E}{(\mathbf{x}_i+\mathbf{x}_j-2\mathbf{x}_i\odot\mathbf{x}_j)}}+\beta\sum_{k=1}^{K} {({|P_k|-\frac{|V|}{K}})}^2+\phi(\mathbf{x})
\label{equ:GP-OHQUBO}
\end{equation}
where ${\mathbf{x}_i},{\mathbf{x}_j}$ are $K$-dimensional one-hot encoded decision vectors, $\alpha$ and $\beta$ are hyperparameters, $K$ is the predefined number of partitioned blocks, and $|P_k|$ denotes the number of vertices in partitioned block $P_k$.
The OH-QUBO term $\sum{(\mathbf{x}_i+\mathbf{x}_j-2\mathbf{x}_i\odot\mathbf{x}_j)}$ represents the number of cut edges. $\sum{(\mathbf{x}_i+\mathbf{x}_j-2\mathbf{x}_i\odot\mathbf{x}_j)}=0$ if $\mathbf{x}_i$ and $\mathbf{x}_j$ are assigned the same color, or otherwise it equals to 2.
$\sum_{k=1}^{K} {({|P_k|-\frac{|V|}{K}})}^2$ ensures the balancedness.

\textbf{Hypergraph Partitioning.} Given a hypergraph $G=(V,E)$, the hypergraph partitioning problems can be expressed as:
\begin{equation}
\mathrm{min} \quad O_{\mathrm{HP}}=\alpha\sum{\sum_{e_j\in E}\frac{1}{|e_j|}{(\sum_{i=1}^{|e_j|}{\mathbf{x}_{ji}}-|e_j|{\mathbf{x}_{j1}}\odot\cdots\odot{\mathbf{x}_{j|e_j|}})}}+\beta\sum_{k=1}^{K} {({|P_k|-\frac{|V|}{K}})}^2+\phi(\mathbf{x})
\label{equ:HP-OHPUBO}
\end{equation}
where ${\mathbf{x}_{ji}}$ are $k$-dimensional one-hot encoded decision vectors of vertices $v_{ji}\in e_j$. $K$ is the predefined number of partitioned blocks.
$\alpha$ and $\beta$ are hyperparameters.
The OH-PUBO term $\sum{\sum_{e_j\in E}\frac{1}{|e_j|}{(\sum_{i=1}^{|e_j|}{\mathbf{x}_{ji}}-|e_j|{\mathbf{x}_{j1}}\odot\cdots\odot{\mathbf{x}_{j|e_j|}})}}$ represents the number of cut edges.

\subsection{Neural Network-based Optimizer}
\label{sec:NNarchitecture}
Deep $k$-grouping solves OH-QUBO or OH-PUBO problems on graphs or hypergraphs using GNNs or hypergraph neural networks (HyperGNNs) as illustrated in Fig.~\ref{fig:framework}.

\textbf{Model Input:} For a graph $G=(V,E)$, the GNN input consists of the adjacency matrix $A$ and randomly initialized node features $X^{(0)} \in \mathbb{R}^{|V|\times d^{(0)}}$.
For a hypergraph $G=(V,E)$, HyperGNNs take the incidence matrix $H$ and the same randomly initialized $X^{(0)}$ as input.

\textbf{Model Architecture:} Typically, the neural networks combine GNN (or HyperGNN) layers with fully connected (FC) layers, with architectures intrinsically adapted to the problem.
For instance, graph (hypergraph) partitioning necessitates the capture of global structural information, thereby requiring a deep GNN (or HyperGNN) model (4-8 layers) supplemented by FC layers. In contrast, the graph coloring problem emphasizes local information, thus demanding a shallow model (2-3 layers) comprised solely of GNN layers.

\textbf{Model Output:} The neural networks apply the softmax function to produce the relaxation of one-hot decision variables, represented as a matrix $X\in\mathbb{R}^{|V|\times k}$.
The $i$-th row of $X$, denoted by $\mathbf{x}_i\in [0,1]^k$, corresponds to the assignment probability vector of vertex $v_i$ over $k$ groups.

More concretely, the neural network model operates as follows:
\begin{equation}
X=\mathrm{Softmax}(\mathrm{FC}(\mathrm{GNN}(A,X^{(0)})))\quad \mathrm{or}\quad X=\mathrm{Softmax}(\mathrm{FC}(\mathrm{HyperGNN}(H,X^{(0)})))
\label{equ:NNmodel}
\end{equation}
where GNN or HyperGNN is a multi-layer graph or hypergraph convolutional network.

\textbf{Loss Function:}
The neural network-based optimizer aims to minimize the OH-QUBO or OH-PUBO objective function.
To achieve this, Deep $k$-grouping employs the softmax-relaxation of OH-QUBO or OH-PUBO objective function as a differentiable loss function $L(X)$, with respect to $X$, the output of the neural network model.
We aim to find the optimal solution $X^{s}=\mathrm{argmin}L(X)$.

\textbf{Training to Optimize:} On this basis, Deep $k$-grouping optimizes the loss function through unsupervised training.
Meanwhile, the group assignment matrix $X$ will converge to approximate solutions $X^{s}$ of OH-QUBO or OH-PUBO problems.

\subsection{Gini Coefficient-based Annealing}
\label{sec:annealing}
Neural network-based unsupervised CO solvers face two issues: \textbf{(\romannumeral1)} Neural networks are prone to getting stuck in local optima; \textbf{(\romannumeral2)} The differentiable relaxation of discrete decision variables leads to soft solutions of continuous values.
To address these issues, CRA-PI-GNN~\cite{ichikawa2024controlling} introduces a continuous relaxation annealing strategy.
It leverages the penalty term $\gamma \sum_{i=1}^{n}(1-(2x_i-1)^{\alpha})$ to avoid local optima and enforce the discreteness of solutions, where $x_i\in [0,1]$ are relaxed variables, $\gamma$ controls the penalty strength and $\alpha$ is an even integer.

Inspired by CRA-PI-GNN, we explore the continuous relaxation annealing strategy for the softmax relaxation of one-hot variables $\mathbf{x}_i\in [0,1]^k$.
Let $\mathbf{x}_i = (x_{i1}, x_{i2}, \cdots, x_{ik})$ be a probability vector, where $\sum_{j=1}^k x_{ij} = 1$ and $x_{ij} \geq 0$.
An intuitive approach is to use Shannon entropy~\cite{pereyra2017regularizing} term $\gamma\sum_{i=1}^{n}\sum_{j=1}^{k}-x_{ij}\mathrm{log}x_{ij}$ as the penalty.
However, Shannon entropy function is infeasible for neural networks since the gradient diverges to $\pm\infty$ at $0$ or $1$.
To address this issue, we need to find an alternative function that exhibits the same trend of variation as the Shannon entropy.
To this end, we employ Gini coefficient-based penalty term:
\begin{equation}
\phi(\mathbf{x})=\gamma\sum_{i=1}^{n}(1-\sum_{j=1}^{k}x_{ij}^2)
\label{equ:Gini}
\end{equation}
For a one-hot vector $\mathbf{x}_i$, the value of Gini coefficient $1-\sum_{j=1}^{k}x_{ij}^2=0$, or otherwise when $\mathbf{x}_i$ is uniform ($x_{ij}=\frac{1}{k}$), the maximum value of Gini coefficient is $1-\frac{1}{k}$.
The annealing strategy enhances the learning process by gradually annealing the parameter $\gamma$: \textbf{(\romannumeral1)} Initially, $\gamma$ is set to a negative value ($\gamma<0$) to smooth the model, thereby facilitating broad search of the solution space and avoiding local optima; \textbf{(\romannumeral2)} Subsequently, the value of $\gamma$ is gradually increased (until $\gamma>0$) to enforce the discreteness of solutions.

\section{Experiments}
\label{sec:experiments}
\textbf{Datasets and Baseline Methods.} We evaluate the performance of Deep $k$-grouping on synthetic and real-world graph and hypergraph datasets (refer to Tab.~\ref{tab:datasets_graph} in Appendix~\ref{app:datasets}).
Baseline methods include \textbf{(\romannumeral1)} IP solvers (e.g., SCIP~\cite{MaherMiltenbergerPedrosoRehfeldtSchwarzSerrano2016} and Tabu~\cite{glover1998tabu}); \textbf{(\romannumeral2)} Neural network solvers (e.g., GAP~\cite{nazi2019deep} for graph partitioning); \textbf{(\romannumeral3)} Problem-specific heuristics (e.g., hMETIS~\cite{karypis1997multilevel} and KaHyPar~\cite{gottesburen2024scalable} for graph/hypergraph partitioning).

\textbf{Implementation.}
The GPU used in the experiment is an NVIDIA GeForce RTX 3090 with 24 G of memory.
We implement GNNs and HyperGNNs with DHG library~\cite{gao2022hgnn+,feng2019hypergraph}.
\subsection{Graph/Hypergraph Coloring}
\begin{table*}[t]
\setlength{\tabcolsep}{2pt}
\renewcommand{\arraystretch}{0.2}
\caption{Experimental Results for Graph Coloring.}
\small
\centering
\begin{tabular}{lllllllll}
\toprule
Method&
BAT&
EAT&
UAT&
DBLP&
CiteSeer&
AmzPhoto&
AmzPc\\
\midrule
SCIP&
18 (2 s)&
29 (9 s)&
64 (25 s)&
8 (16 s)&
8 (22 s)&
-&
-\\
&
\underline{16} (16 s)&
\underline{25} (12 s)&
\underline{60} (32 s)&
7 (37 s)&
7 (66 s)&
-&
-\\
Tabu&
25 (17 s)&
53 (59 s)&
79 (898 s)&
12 (123 s)&
15 (195 s)&
146 (2,770 s)&
-\\
Ours&
18 (2 s)&
29 (2 s)&
65 (8 s)&
\textbf{7} (4 s)&
\textbf{7} (20 s)&
\textbf{48} (90 s)&
\textbf{52} (263 s)\\
\bottomrule
\end{tabular}
\label{tab:graphcoloring}
\end{table*}
\begin{table*}[t]
\setlength{\tabcolsep}{2pt}
\renewcommand{\arraystretch}{0.2}
\caption{Experimental Results for Hypergraph Proper Coloring.}
\small
\centering
\begin{tabular}{llllll}
\toprule
Method&
Primary&
High&
Cora&
PubMed&
Cooking200\\
\midrule
SCIP&
29 (12 s)&
20 (424 s)&
4 (13 s)&
5 (15 s)&
3 (36 s)\\
&
\underline{27} (557 s)&
20 (424 s)&
\underline{3} (30 s)&
\underline{4} (27 s)&
\underline{2} (851 s)\\
Tabu&
41 (3,378 s)&
26 (2,124 s)&
5 (102 s)&
7 (254 s)&
4 (1,654 s)\\
Ours&
35 (5 s)&
\textbf{20} (6 s)&
4 (8 s)&
5 (7 s)&
3 (6 s)\\
\bottomrule
\end{tabular}
\label{tab:hypergraphcoloring}
\end{table*}
We use real-world datasets (Tab.~\ref{tab:graphcoloring} and Tab.~\ref{tab:hypergraphcoloring}) to evaluate the performance of Deep $k$-grouping on graph coloring and hypergraph proper coloring.
We use $2$-layer GraphSAGE~\cite{hamilton2017inductive} as the learning model, since various GNNs and HyperGNNs have been examined and GraphSAGE always achieves the best performance on either graphs or hypergraphs.
To adapt GraphSAGE for vertex embedding tasks in hypergraph proper coloring, the hypergraphs are transformed into graphs through clique expansion.
We apply max pooling to generate decision variables for colors $\mathbf{y}=(y_1,\cdots,y_{K_{max}})$ in Eq.~\ref{equ:GC-OHQUBO} or Eq.~\ref{equ:HSC-OHQUBO} as follows: $\mathbf{y}=\mathrm{Sigmoid}(\mathrm{MP}(\mathrm{GraphSAGE}(A,X^{(0)})))$, where $\mathrm{MP}$ is the max pooling layer.
The values of penalty parameters $\lambda_1$, $\lambda_2$ are initialized to 0 and gradually increased during training to satisfy the constraints.
The learning rate $\eta$ is set to $1^{-4}$.

As depicted in Tab.~\ref{tab:graphcoloring} and Tab.~\ref{tab:hypergraphcoloring},
the chromatic number and the initial time to find a corresponding solution (e.g., 18 (2 s)) are depicted in the tables.
SCIP outperforms Deep $k$-grouping and Tabu on small datasets such as BAT, EAT, and UAT.
However, we ran SCIP on large-scale datasets including AmzPc and AmzPhoto for 5 hours but cannot find any feasible solution.
That is, Deep $k$-grouping achieves superior performance to SCIP and Tabu for large-scale graph coloring problems.
For hypergraph proper coloring, we observe that while Deep $k$-coloring can color hypergraphs faster, yet it struggles to find the optimal chromatic number as SCIP does.
Notably, SCIP's significant performance is achieved through novel objective functions we proposed in Sec.~\ref{sec:proglem_instances}.
\subsection{Graph/Hypergraph Partitioning}
\begin{table*}[h]
\setlength{\tabcolsep}{2pt}
\renewcommand{\arraystretch}{0.2}
\caption{Performance comparison of graph/hypergraph partitioning methods including Deep $k$-grouping, GAP (neural network-based), hMETIS, and KaHyPar (problem-specific multi-level heuristics). BAT, EAT, UAT, DBLP, CiteSeer are graphs. Primary, High, Cora, PubMed, and Cooking200 are hypergraphs. Three metrics—$C\ (B_1,B_2)$ have been illustrated in the table, where $C$ is the number of cut edges, $B_1=\frac{\mathrm{max}(S)}{\mathrm{mean}(S)}-1$ and $B_2=\sqrt{\frac{1}{k}\sum_{i=1}^{k}{(|P_i|-\mathrm{mean}(S)})^2}$ are two types of balancedness (the smaller the value of $B1$ and $B2$, the better the balancedness), and $S=\{|P_1|,|P_2|,...,|P_k|\}$ is the set of partition block sizes.}
\small
\centering
\begin{tabular}{lllllll}
\toprule
Dataset&
Method&
$k=2$ &
$k=3$ &
$k=4$ &
$k=5$ &
$k=6$ \\
\midrule
BAT&
hMETIS&
192 (2.3\%,1.5)&
368 (5.3\%,2.1)&
465 (6.9\%,2.3)&
528 (11\%,1.9)&
593 (9.9\%,1.3)\\
&
GAP&
192 (2.3\%,1.5)&
362 (5.3\%,1.7)&
456 (6.9\%,1.3)&
526 (6.9\%,1.0)&
569 (9.9\%,1.1)\\
&
SCIP&
198 (0\%,0)&
369 (0\%,0)&
491 (0\%,0)&
556 (0\%,0)&
589 (0\%,0)\\
&
Ours&
\textbf{191} (2.3\%,1.5)&
\textbf{346} (5.3\%,1.7)&
\textbf{446} (6.9\%,1.3)&
\textbf{514} (6.9\%,1.2)&
\textbf{567} (9.9\%,1.1)\\
\midrule
EAT&
hMETIS&
\underline{946} (2.3\%,4.5)&
2,034 (3.8\%,3.6)&
2,565 (5.3\%,5.3)&
2,990 (6.5\%,4.5)&
2,983 (46\%,22)\\
&
GAP&
\underline{946} (2.3\%,4.5)&
2,047 (3.8\%,3.6)&
2,626 (6.3\%,3.7)&
2,917 (6.5\%,3.5)&
2,913 (50\%,15)\\
&
SCIP&
996 (0\%,0)&
2,224 (0\%,0)&
3,065 (0\%,0)&
3,863 (0\%,0)&
4,965 (0\%,0)\\
&
Ours&
\textbf{946} (2.3\%,4.5)&
\textbf{2,015} (3.8\%,3.6)&
\textbf{2,554} (5.3\%,3.9)&
\textbf{2,915} (6.5\%,3.2)&
\textbf{2,871} (43\%,13)\\
\midrule
UAT&
hMETIS&
411 (2.0\%,12)&
1,078 (3.9\%,11)&
1,783 (4.2\%,9.0)&
2,568 (5.5\%,11)&
2,877 (46\%,61)\\
&
GAP&
410 (2.2\%,13)&
1,138 (4.1\%,12)&
1,842 (3.2\%,5.5)&
2,585 (5.0\%,6.0)&
2,820 (21\%,18)\\
&
SCIP&
424 (0\%,0)&
1,488 (0\%,0)&
2,643 (0\%,0)&
8,997 (0\%,0)&
11,159 (0\%,0)\\
&
Ours&
\textbf{407} (2.0\%,12)&
\textbf{1,076} (3.9\%,11)&
\textbf{1,755} (3.2\%,5.5)&
\textbf{2,499} (4.6\%,5.5)&
\textbf{2,719} (20\%,18)\\
\midrule
DBLP&
hMETIS&
84 (0\%,0)&
136 (0\%,0)&
166 (0\%,0)&
186 (0\%,0)&
223 (0\%,0)\\
&
GAP&
198 (0\%,0)&
322 (0\%,0)&
332 (0\%,0)&
432 (0.2\%,0.7)&
399 (8.6\%,19)\\
&
SCIP&
\underline{38} (0\%,0)&
\underline{83} (0\%,0)&
222 (0\%,0)&
1,248 (0\%,0)&
515 (0\%,0)\\
&
Ours&
74 (0\%,0)&
113 (0\%,0)&
\textbf{156} (0\%,0)&
\textbf{168} (0\%,0)&
\textbf{213} (0.8\%,2.5)\\
\midrule
CiteSeer&
hMETIS&
27 (0\%,0)&
70 (0\%,0)&
101 (0.2\%,0.8)&
145 (0\%,0)&
152 (0\%,0)\\
&
GAP&
193 (0\%,0)&
430 (0\%,0)&
426 (0.2\%,0.8)&
440 (0.3\%,1.7)&
536 (0\%,0)\\
&
SCIP&
\underline{25} (0\%,0)&
\underline{27} (0\%,0)&
\underline{49} (0\%,0)&
140 (0\%,0)&
1258  (0\%,0)\\
&
Ours&
39 (0\%,0)&
62 (0\%,0)&
93 (0\%,0)&
\textbf{118} (0\%,0)&
\textbf{142} (0.3\%,0.8)\\
\midrule

Primary&
KaHyPar&
\underline{2,582} (0.8\%,1)&
\underline{4,070} (0\%,0)&
\underline{5,072} (0\%,0)&
5,190 (0\%,0)&
6,576 (0\%,0)\\
&
SCIP&
3,614 (0\%,0)&
7,792 (0\%,0)&
8,758 (0\%,0)&
11,323 (0\%,0)&
11,420 (0\%,0)\\
&
Ours&
\textbf{2,582} (0.8\%,1)&
\textbf{4,070} (0\%,0)&
\textbf{5,072} (0\%,0)&
\textbf{5,181} (0\%,0)&
\textbf{6,484} (0\%,0)\\
\midrule
High&
KaHyPar&
793 (1.0\%,2.5)&
828 (1.0\%,0.8)&
1,530 (0\%,0)&
1,916 (0\%,0)&
2,151 (0\%,0)\\
&
SCIP&
905 (0\%,0)&
1,520 (0\%,0)&
6,127 (0\%,0)&
6,389 (0\%,0)&
6,558 (0\%,0)\\
&
Ours&
\textbf{784} (1.5\%,2.5)&
\textbf{812} (2.8\%,2.2)&
\textbf{1,523} (0\%,0)&
\textbf{1,903} (0\%,0)&
\textbf{2,149} (0\%,0)\\
\midrule
Cora&
KaHyPar&
123 (0\%,0)&
172 (0.4\%,1.7)&
201 (0\%,0)&
237 (0.8\%,2.6)&
270 (0.6\%,2.2)\\
&
SCIP&
\underline{116} (0\%,0)&
\underline{151} (0\%,0)&
210 (0\%,0)&
\underline{232} (0\%,0)&
324 (0\%,0)\\
&
Ours&
122 (0\%,0)&
169 (0\%,0)&
\textbf{200} (0\%,0) &
234 (0\%,0)&
\textbf{261} (0\%,0)\\
\midrule
PubMed&
KaHyPar&
\underline{814} (0\%,0)&
1,253 (0\%,0)&
1,476 (0.6\%,9.3)&
1,864 (0.9\%,12)&
2,083 (1.1\%, 15)\\
&
SCIP&
911 (0\%,0)&
5,463 (0\%,0)&
6,388 (0\%,0)&
5,887 (0\%,0)&
-\\
&
Ours&
844 (0\%,0)&
\textbf{1,157} (0\%,0)&
\textbf{1,455} (0.3\%,3.1)&
\textbf{1,855} (0.8\%,9.5)&
\textbf{2,048} (0\%,0)\\
\midrule
Cooking&
KaHyPar&
1,093 (0\%,0)&
\underline{1,304} (0\%,0)&
1,454 (0\%,0)&
1,507 (0\%,1.5)&
1,560 (0.7\%,7.9)\\
200&
SCIP&
\underline{1,073} (0\%,0)&
1,978 (0\%,0)&
2,395 (0\%,0)&
2,022 (0\%,0)&
2,036 (0\%,0)\\
&
Ours&
1,283 (0\%,0)&
1,305 (0\%,0)&
\textbf{1,406} (0\%,0)&
\textbf{1,439} (0\%,0.8)&
\textbf{1,511} (0\%,0)\\
\bottomrule
\end{tabular}
\label{tab:graphhypergraphpartitioning}
\end{table*}
\begin{wrapfigure}[15]{r}{4.5cm}
  \includegraphics[width=4.5cm]{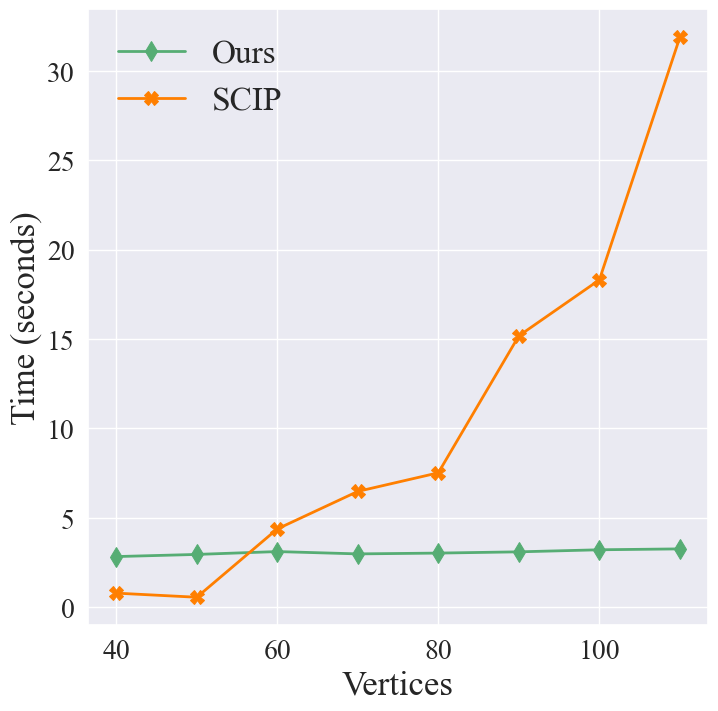}
  \caption{Comparison of the runtime between SCIP and Deep $k$-grouping.}
  \label{fig:runtimeDeepKSCIP}
\end{wrapfigure}
We evaluate the performance of Deep $k$-grouping on graph/hypergraph partitioning as depicted in Tab.~\ref{tab:graphhypergraphpartitioning}.
Best performed GAT~\cite{velivckovic2018graph} with $2$ or $3$ layers is applied to construct Deep $k$-grouping for graph partitioning.
For hypergraph partitioning, Deep $k$-grouping achieves best solutions with $4\sim 8$-layer HGNN+~\cite{gao2022hgnn+} incorporating FC layers.
$\frac{\alpha}{\beta}$ is assigned a fixed value and the learning rate $\eta$ is set to $1^{-4}$.
The runtime of SCIP was limited to 1 hour.
As the evaluation results shown in Tab.~\ref{tab:graphhypergraphpartitioning}, on small datasets, SCIP can find the optimal solution. However, as the data scale and the value of $k$ increase, the solution space grows exponentially, making it challenging for SCIP to approximate the optimal solution. In such scenarios, Deep $k$-grouping outperforms other methods.
\subsection{Runtime Comparison with SCIP}
We conduct additional experiments to comprehensively evaluate the solving time differences between Deep $k$-grouping and SCIP.
The single-objective max-cut problem on graphs (refer to Appendix~\ref{app:maxcut}) is adopted to exclude the effects of extraneous optimization objectives such as balancedness.

The runtime of SCIP and Deep $k$-grouping has been tested when vertex counts $|V|$ of graphs $G=(V,E)$ are ranging from $40$ to $110$, and $|E|=4|V|$.
SCIP was terminated immediately when it found a feasible solution.
We trained Deep $k$-grouping until it converged.
As illustrated in Fig.~\ref{fig:runtimeDeepKSCIP}, Deep $k$-grouping's runtime remains stable while SCIP’s runtime grows exponentially with graph size.
This further explains why Deep $k$-grouping achieves better performance on large-scale datasets.
\subsection{Ablation Study on the Annealing Strategy}
We validate the effectiveness of Deep $k$-grouping with or without Gini coefficient-based annealing strategy.

\begin{figure}[t]
\centering
\begin{minipage}[b]{0.32\textwidth}
\includegraphics[width=\linewidth]{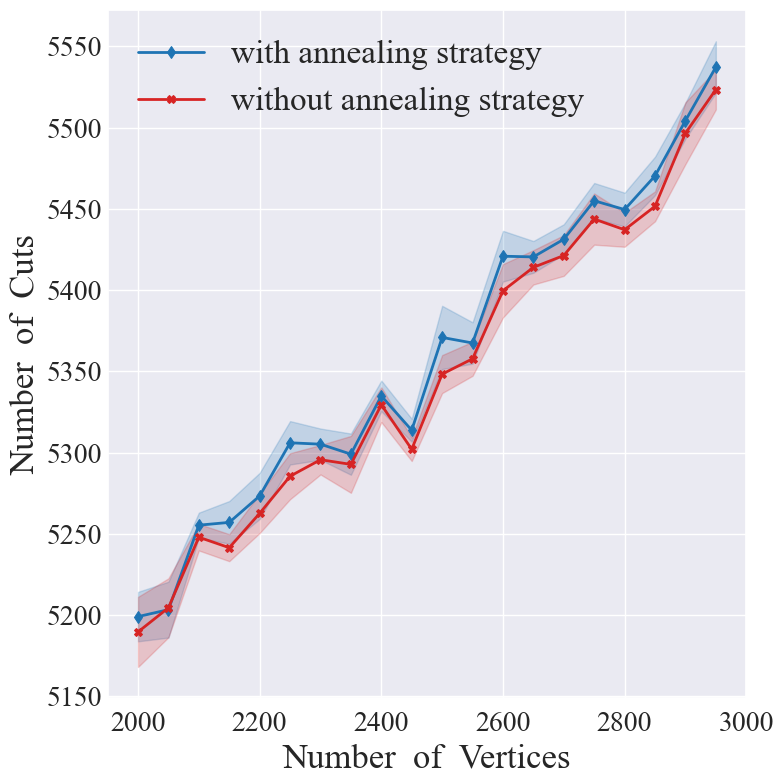} 
\caption{Quality of solutions of the max-cut problem with or without Gini coefficient-based annealing strategy across various graph sizes.}
\label{fig:ginisolutionqual}
\end{minipage}
\hfill
\begin{minipage}[b]{0.32\textwidth}
    \includegraphics[width=\linewidth]{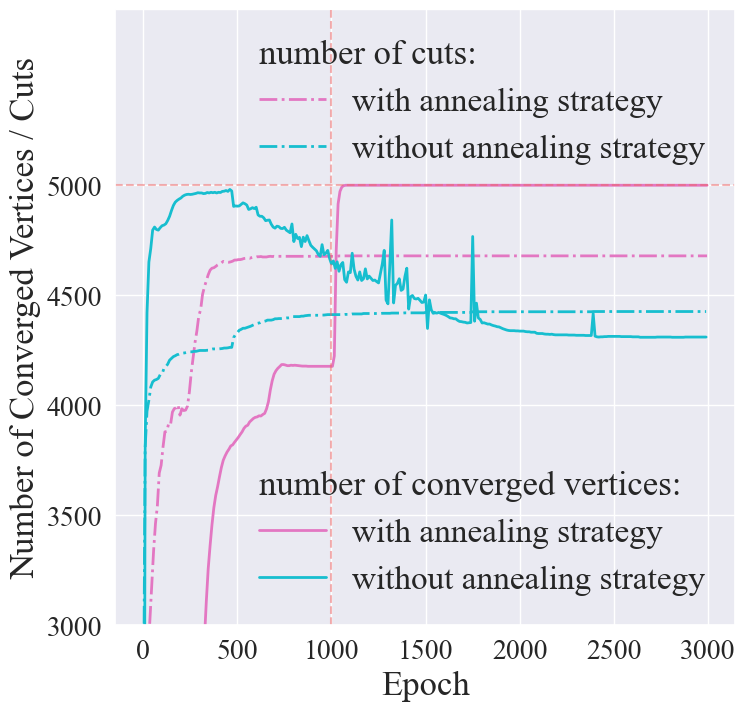}
    \caption{Quality and discreteness of solutions of the max-cut problem with or without Gini coefficient-based annealing strategy.}
    \label{fig:ginicompare2}
\end{minipage}
\hfill
\begin{minipage}[b]{0.32\textwidth}
    \includegraphics[width=\linewidth]{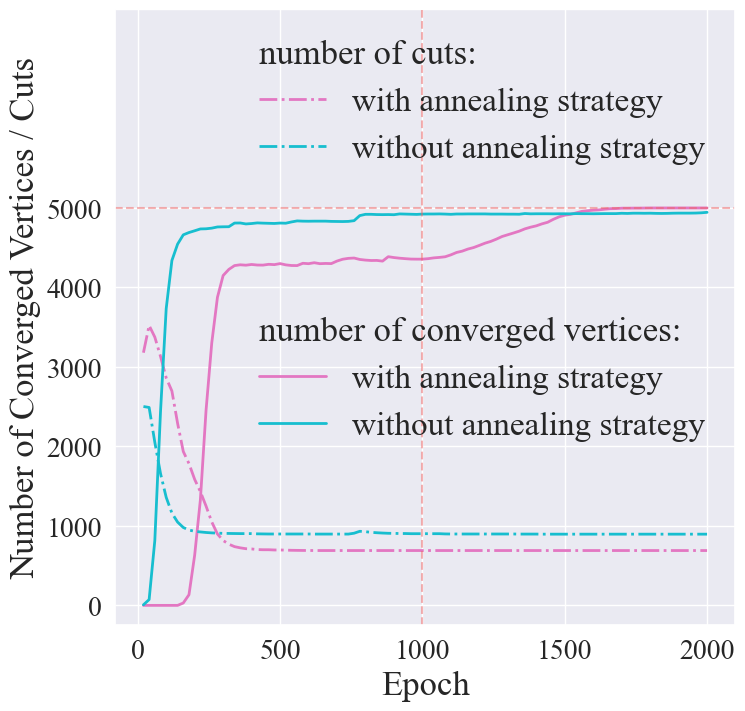}
    \caption{Quality and discreteness of solutions of the graph partitioning problem with or without Gini coefficient-based annealing strategy.}
    \label{fig:ginicompare3}
\end{minipage}
\end{figure}

As illustrated in Fig.~\ref{fig:ginisolutionqual}, we first evaluated the quality of solutions of the max-cut problem across different graph scales.
The experiments tested the performance of the Gini coefficient-based annealing strategy on graphs with $2,000\sim3,000$ vertices, generating 50 random graphs (where $|V|=|E|$) each time and averaging the cut numbers.
Experimental results in Fig.~\ref{fig:ginisolutionqual} demonstrate that the Gini coefficient-based annealing strategy effectively enhances solution quality, demonstrating its ability to prevent the model from getting trapped in local optima.

We also conducted experiments to verify that Gini coefficient-based annealing enforces the discreteness of solutions.
The experiments were conducted on a graph with 5,000 vertices and 5,000 edges. It measured the quality of solutions and their convergence behavior on the max-cut problem with increasing training epochs.
Initially, the penalty strength $\gamma$ was set to $-2.5$ and gradually increased its value during training.
It reached 0 after 1000 epochs and continued to increase thereafter.
As shown in Fig.~\ref{fig:ginicompare2}, the Gini coefficient-based annealing strategy ensures sustained acquisition of higher-quality solutions while guaranteeing complete convergence to discrete values. In contrast, when this annealing strategy is disabled, a growing proportion of nodes exhibit failed convergence to discrete solutions as training epochs increase.

We further evaluated the Gini coefficient-based annealing strategy for the graph partitioning problem ($k=4$).
The experiments were conducted on a graph comprising 5,000 vertices and 5,000 edges.
The penalty strength $\gamma$ was set to $-0.25$ and reached 0 after 1000 epochs and continued to increase thereafter.
We monitored the cut number and discreteness of solutions with increasing training epochs.
As illustrated in Fig.~\ref{fig:ginicompare3}, negative $\gamma$ values enable the model to escape local optima, thereby discovering higher-quality solutions.
Conversely, positive $\gamma$ guarantees convergence of all vertices states to discrete assignments, which rigorously enforces the constraints.
Conversely, when parameter $\gamma$ is set to positive values, the system exhibits accelerated convergence dynamics, where all variables rigorously converge to discrete states eventually.

\section{Conclusions}
In this work, we propose Deep $k$-grouping, an unsupervised neural network solver for solving $k$-grouping CO problems on graphs and hypergraphs.

\textbf{Algorithm Design.} We propose OH-QUBO and OH-PUBO formalisms for modeling $k$-grouping CO problems.
On this basis, we propose a GPU-accelerated and Gini coefficient-based annealing enhanced training pipeline to solve $k$-grouping CO problems in an end-to-end manner.

\textbf{Extensive Experiments \& Empirical Findings.}
We propose novel cost functions for graph/hypergraph coloring/partitioning problems.
Although exact solvers like SCIP can exhaustively search for globally optimal solutions for small-scale datasets, experimental results have demonstrated the performance of Deep $k$-grouping on large-scale optimization problems.
Deep $k$-grouping outperforms existing approximate solvers, including neural network solvers and problem-specific heuristics.
Neural solvers such as Deep $k$-grouping establish new state-of-the-art performance and hold promise in tackling real-world CO challenges.

\bibliographystyle{unsrt}
\bibliography{citation}
\newpage
\appendix
\onecolumn

\section{Related works}
\label{Related_Works}
PI-GNN~\cite{schuetz2022combinatorial} introduced the QUBO formulation from quantum computing into unsupervised neural CO. Building upon PI-GNN, a series of enhanced methods have been proposed, such as value classification network (VCN)~\cite{chen2024learning} and continuous relaxation annealing (CRA)~\cite{ichikawa2024controlling}. While these advancements demonstrate the potential of unsupervised learning in CO, prior studies primarily validated them on basic CO problems, such as MIS and max-cut.

Researchers have also formulated graph matching~\cite{wang2021neural} as the quadratic assignment problem (QAP, a classic QUBO problem) and addressed it through unsupervised neural networks.
Owing to its critical applications in computer vision, graph matching has been extensively investigated~\cite{xie2024contrastive,zhou2023improving,tourani2024discrete}.
In the field of $k$-grouping CO, GNN-based unsupervised learning models include GAP~\cite{nazi2019deep} and Deep Modularity Networks (DMN)~\cite{tsitsulin2023graph} for graph partitioning and clustering.
Most of these neural solvers are problem-specific.

In summary, prior methods primarily focus on CO problems with either binary variables for $2$-grouping problems or quadratic-cost CO problems on graphs.
In this work, we extend unsupervised CO frameworks to industrially critical yet underexplored problems, such as hypergraph partitioning~\cite{bustany2022specpart,bustany2023k,sybrandt2022hypergraph}—a key challenge in VLSI design and distributed computing.

\section{The MIS problem}
\label{app:MISDef}
The MIS problem for an ordinary graph $G=(V,E)$ can be formulated by counting the number of marked (colored by 1) vertices belonging to the independent set and adding a penalty when two vertices are connected by an edge.

\textbf{IP Form.} Given a graph $G=(V,E)$, the MIS problem can be formulated as the following IP form:
\begin{align}
\mathrm{min} &\quad -\sum_{i\in V}x_i
\label{equ:MIS-IP1}
\\ \mathrm{s.t.} &\quad x_ix_j=0 \quad \mathrm{for}\ \mathrm{all}\quad (i,j)\in E
\label{equ:MIS-IP2}
\\ &\quad x_i\in\{0,1\}
\label{equ:MIS-IP3}
\end{align}
\textbf{QUBO Form.} The MIS problem can be formulated as the QUBO form by using penalty method:
\begin{equation}
    \mathrm{min} \quad -\sum_{i\in V}x_i+P\sum_{(i,j)\in E}{x_ix_j}
\label{equ:QUBO-MIS}
\end{equation}
where $P$ is the penalty parameter and $x_i\in\{0,1\}$. $P>0$ enforces the constraint that $x_i$ and $x_j$ cannot both be equal to 1 simultaneously.

\textbf{Unsupervised Neural Network-based Optimizer.} As illustrated in Fig.~\ref{fig:framework}, initially, the MIS problem is phrased in QUBO form by constructing the matrix $Q= \begin{cases} 
P & \text{if } (i,j)\in E \\
-1 & \text{if } i=j \\
0 & \text{otherwise} \ (\text{feasible})
\end{cases}$. Subsequently, the random parameter vectors are then fed into graph neural networks (GNN) as node features, while relaxed decision variables $x_i\in [0,1]$ are generated through sigmoid function. Through training to minimized the loss function $\mathbf{x}^TQ\mathbf{x}$, decision variables $x_i$ will eventually converge to solutions that are close to discrete values $\{0,1\}$.
\section{The max-cut problem}
\label{app:maxcut}
The max-cut problem of a graph $G=(V,E)$ involves partitioning the vertex set into two disjoint subsets such that the number of edges crossing the partitioned blocks is maximized.

\textbf{QUBO Form.} The max-cut problem can be formulated as the QUBO form:
\begin{equation}
    \mathrm{min} \quad {\sum_{(i,j)\in {E}}(2x_ix_j-x_i-x_j)}
\label{equ:QUBOmaxcut}
\end{equation}
where $x_i\in \{0,1\}$.

\textbf{OH-QUBO Form.} The max-cut problem can be formulated as the OH-QUBO form:
\begin{equation}
    \mathrm{min} \quad {\sum{X^TQ\odot X^T}}
\label{equ:OHQUBOmaxcut}
\end{equation}
where $\mathbf{x}_i\in \{0,1\}^2$ are $2$-dimensional one-hot vectors, and the matrix 
$[\mathbf{x}_1,\mathbf{x}_2,...,\mathbf{x}_{|V|}]$ is denoted by $X\in \mathbb{R}^{|V|\times 2}$.
 The OH-QUBO matrix $Q$=
 $\begin{cases} 
2 & \text{if } (i,j)\in E \\
-2 & \text{if } i=j \\
0 & \text{otherwise}
\end{cases}$.

\section{Proof of GPU-accelerated OH-QUBO cost function in Eq.~\ref{equ:GPUOHQUBO}}
\label{app:proof-Eq3}
\begin{figure*}[t]
\centering
\includegraphics[width=1\columnwidth]{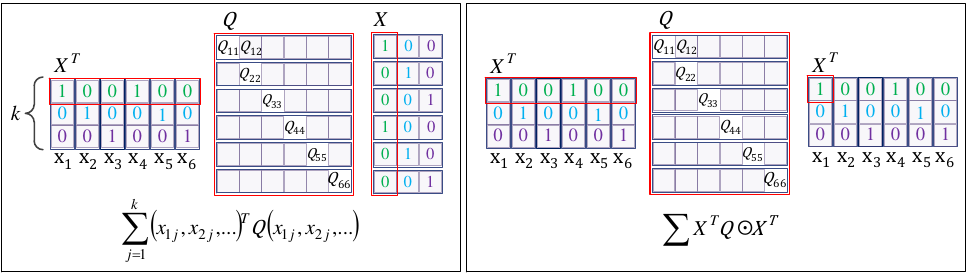} 
\caption{On the proof of Eq.~\ref{equ:GPUOHQUBO}.}
\label{fig:OHQUBODef}
\end{figure*}
The cost function of OH-QUBO can be expressed as:
\begin{equation}
O_{\mathrm{OH-QUBO}}=\sum_{i=1}^{|V|}\sum_{j=1}^{|V|}{Q_{ij}\mathrm{x}_i\odot\mathrm{x}_j}=\sum_{j=1}^{k}{(x_{1j},x_{2j},...)^TQ(x_{1j},x_{2j},...)}
\label{equ:OHQUBOGPUproof}
\end{equation}
where $\mathbf{x}_i$ are $k$-dimensional one-hot vectors.
$x_{ij}\in \{0,1\}$ is the $j$-th element of the one-hot vector $\mathbf{x}_i$.
$Q$ is the OH-QUBO matrix.
From Fig.~\ref{fig:OHQUBODef}, we can see that these two methods $O_{\mathrm{OH-QUBO}}=\sum_{j=1}^{k}{(x_{1,j},x_{2,j},...)^TQ(x_{1,j},x_{2,j},...)}$ and $O_{\mathrm{OH-QUBO}}=\sum{X^TQ\odot X^T}$ (Eq.~\ref{equ:GPUOHQUBO}) are equivalent.
\begin{figure*}[t]
\centering
\includegraphics[width=0.95\columnwidth]{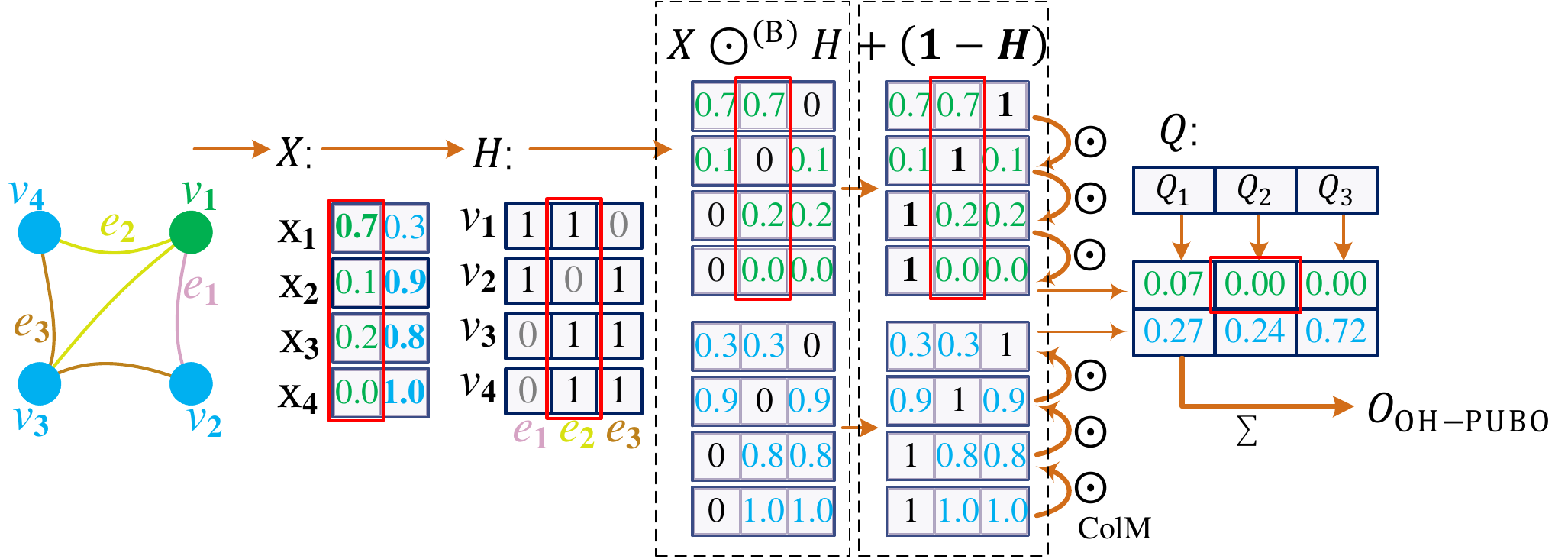} 
\caption{A toy example for Eq.~\ref{equ:OHPUBOTheorem}.}
\label{fig:OHQUBOEXpApp}
\end{figure*}

Note that Eq.~\ref{equ:GPUOHQUBO} holds only if $\mathbf{x}_i$ are discrete one-hot vectors.
When relaxing $O_{\mathrm{OH-QUBO}}$ into a differentiable loss function via softmax function, we have $\mathbf{x}_i\odot \mathbf{x}_i\neq \mathbf{x}_i$ (e.g., $0.5^2\neq 0.5$).
Thus the linear terms and quadratic terms must be handled separately as follows:
\begin{equation}
O_{\mathrm{OH-QUBO}}=\sum_{\mathrm{reduce-sum}}{X^T\text{diag}(Q)}+\sum_{\mathrm{reduce-sum}}{X^T(Q-\text{diag}(Q))\odot X^T}
\label{equ:OHQUBOextend}
\end{equation}
where $\mathrm{diag}(Q)$ denotes the diagonal matrix of $Q$.
\section{Proof of GPU-accelerated OH-PUBO cost function in Eq.~\ref{equ:OHPUBOTheorem}}
\label{app:proof-Eq5}
We present a toy example in Fig.~\ref{fig:OHQUBOEXpApp} to explain the GPU-accelerated OH-PUBO cost function in Eq.~\ref{equ:OHPUBOTheorem}, $O_{\mathrm{OH-PUBO}}=\sum{Q\odot ^{(\mathrm{B})}{\mathrm{ColM}(X\odot ^{(\mathrm{B})}H+^{(\mathrm{B})}(1-H), 1)}}$.
As illustrated in Fig.~\ref{fig:OHQUBOEXpApp}, for the hypergraph $G=(V,E)$, there are four vertices $v_1$, $v_2$, $v_3$ and $v_4$, and three hyperedges $e_1=\{v_1,v_2\}$, $e_2=\{v_1,v_3,v_4\}$, $e_3=\{v_2,v_3,v_4\}$.
The OH-PUBO matrix $Q=[Q_1,Q_2,Q_3]$.
Thus we have $O_{\mathrm{OH-PUBO}}=\sum{({Q_1}\mathbf{x}_1\mathbf{x}_2+{Q_2}\mathbf{x}_1\mathbf{x}_3\mathbf{x}_4+{Q_3}\mathbf{x}_2\mathbf{x}_3\mathbf{x}_4})$.
Initially, we have an output of the neural network-based optimizer $X\in\mathbb{R}^{|V|\times k}$ and the incidence matrix $H\in\mathbb{R}^{|V|\times|E|}$.
By using the broadcasting mechanism, we have $X\odot^{(\mathrm{B})}H\in\mathbb{R}^{|V|\times|E|\times k}$.
To perform column-wise multiplication $\mathrm{ColM}$, we fill the null values with $1$ by adding $(1-H)$.
Finally, we broadcast the OH-PUBO matrix $Q$ to terms $\mathbf{x}_1\mathbf{x}_2$, $\mathbf{x}_1\mathbf{x}_3\mathbf{x}_4$, and $\mathbf{x}_2\mathbf{x}_3\mathbf{x}_4$.

\section{Datasets and baseline methods}
\label{app:datasets}
We evaluate the performance of Deep $k$-grouping with real-world and synthetic datasets.
All publicly available and real-world graph/hypergraph datasets are summarized in Tab.~\ref{tab:datasets_graph}.
For synthetic datasets, we use DHG library to generate graphs and hypergraphs.
\begin{table}[h]
\caption{Summary statistics of seven real-world graphs: the number of vertices $|V|$, the number of edges $|E|$. Five hypergraphs: the number of vertices $|V|$, the number of hyperedges $|E|$, the size of the hypergraph $\sum_{e\in E}{|e|}$.}
\centering
\begin{tabular}{ccc|cccc}
\toprule
\textbf{Graphs} &
$|V|$ &
$|E|$ &
\textbf{Hypergraphs} &
$|V|$ &
$|E|$ &
$\sum_{e\in E}{|e|}$ \\
\midrule
BAT &
131 &
1,003 &
Primary &
242 &
12,704 &
30,729 \\
EAT &
399 &
5,993 &
High &
327 &
7,818 &
18,192 \\
UAT &
1,190 &
13,599 &
Cora &
1,330 &
1,413 &
4,370 \\
DBLP &
2,591 &
3,528 &
Pubmed &
3,824 &
7,523 &
33,687 \\
CiteSeer &
3,279 &
4,552 &
Cooking200 &
7,403 &
2,750 &
54,970 \\
AmzPhoto &
7,535 &
119,081 \\
AmzPc &
13,471 &
245,861 \\
\bottomrule
\end{tabular}
\label{tab:datasets_graph}
\end{table}

The baseline methods used in the experiments are summarized as below:

\textbf{SCIP.} A powerful open-souce optimization solver that can handle mixed-integer programming (MIP) and constraint programming. In our experiment we use a SCIP optimization suit called PySCIPOpt~\cite{MaherMiltenbergerPedrosoRehfeldtSchwarzSerrano2016}, which is an interface from Python.

\textbf{Tabu Search.} A classic metaheuristic search method employing local search methods for combinatorial optimization. It was first formalized in 1989~\cite{glover1998tabu}.

\textbf{GAP.} The first GNN-enabled deep learning framework for graph partitioning~\cite{nazi2019deep}.
GAP employs an end-to-end differentiable loss allowed for unsupervised training.

\textbf{hMETIS.} A multi-level graph and hypergraph partitioner. Well-known multi-level schemes consist of three-phases: coarsening, initial partitioning, and refinement.

\textbf{KaHyPar.} A modern multi-level algorithm~\cite{gottesburen2024scalable} with advanced refinement techniques. It outperforms hMETIS in some cases, especially with tight imbalance constraints.

\end{document}